# Brain Model State Space Reconstruction Using an LSTM Neural Network

**Yueyang Liu[1], Artemio Soto-Breceda[2], Yun Zhao[1], Phillipa Karoly[2], Mark J. Cook[3], David B. Grayden[2], Daniel Schmidt[1], Levin Kuhlmann[1]**

**Abstract**

*Objective*

Kalman filtering has previously been applied to track neural model states and parameters, particularly at the scale relevant to electroencephalography (EEG). However, this approach lacks a reliable method to determine the initial filter conditions and assumes that the distribution of states remains Gaussian. This study presents an alternative, data-driven method to track the states and parameters of neural mass models (NMMs) from EEG recordings using deep learning techniques, specifically a Long Short-Term Memory (LSTM) neural network.

*Approach*

An LSTM filter was trained on simulated EEG data generated by a neural mass model using a wide range of parameters. With an appropriately customised loss function, the LSTM filter can learn the behaviour of NMMs. As a result, it can output the state vector and parameters of NMMs given observation data as the input.

*Main Results*

Test results using simulated data yielded correlations with R squared of around 0.99 and verified that the method is robust to noise and can be more accurate than a nonlinear Kalman filter when the initial conditions of the Kalman filter are not accurate. As an example of real-world application, the LSTM filter was also applied to real EEG data that included epileptic seizures, and revealed changes in connectivity strength parameters at the beginnings of seizures.

*Significance*

Tracking the state vector and parameters of mathematical brain models is of great importance in the area of brain modelling, monitoring, imaging and control. This approach has no need to specify the initial state vector and parameters, which is very difficult to do in practice because many of the variables being estimated cannot be measured directly in physiological experiments. This method may be applied using any neural mass model and, therefore, provides a general, novel, efficient approach to estimate brain model variables that are often difficult to measure.

1 Department of Data Science and AI, Faculty of Information Technology,
Monash University, Clayton, Australia
2 Department of Biomedical Engineering, The University of Melbourne,
Parkville, Australia
3 Department of Medicine, St Vincent's Hospital, The University of
Melbourne, Fitzroy, Australia

## 1. Introduction

Understanding the brain is one of the most challenging problems of science, engineering and medicine. Currently, due to limitations of existing brain imaging and neurophysiological techniques, many neurophysiological variables cannot be measured directly (1,2). Moreover, if they can be measured directly, we cannot do so for all neurons in the brain let alone a small brain region. For instance, current techniques relevant to human applications, such as scalp or intracranial electroencephalography (EEG), record the activities of all the neurons of different population types with a heavy weighting towards cortical pyramidal neurons (3). However, EEG does not have the resolution to track single cell activity or connections between neurons or distinct populations. Brain modelling provides an alternative way to understand brain parameters and neural activity indirectly, and can be used to study healthy brain function as well as diseases like epilepsy, Alzheimer's disease or Parkinson's disease (4–8). One of the aims of modelling is to capture essential temporal and/or spatial characteristics of the brain or brain regions of interest, and to understand how they can emerge from underlying neurophysiological variables such as neural membrane potentials, synaptic strengths and time constants.

At the level of EEG modelling, several approaches have been applied using NMMs to infer these crucial variables at the neural population level (9–15). Different methods analyse the EEG in either the time domain or the frequency domain. One of the most common inference frameworks is Dynamic Causal Modelling (DCM), which is usually applied in the frequency domain (14). Frequency-domain inference potentially offers greater stability in the estimates because they are determined over windows of data.

On the other hand, time-domain inference of neurophysiological variables affords the possibility of capturing instantaneous transitions in these variables. This could be used to understand temporal variations in neural processing in the brain or enable timely control interventions (16) such as electrical stimulation to steer the brain away from an epileptic seizure state. These time-resolved variable estimates, in particular model parameters such as population averaged synaptic strengths, can also potentially provide information about the current brain state (e.g., seizure, non-seizure, pre-seizure, short seizure, long seizure). This is because inferred model parameters can be used to determine where the brain dynamics are currently positioned with respect to the model's bifurcation space. Moreover, such information could be used to make predictions, or to detect, different brain states.

Inferring population averaged membrane potentials, synaptic strengths and time constants, either locally or across the brain from a limited number of EEG measurements, is a challenging task with observability issues at play (17).

Depending on the spatial resolution of the population averaging, there are typically fewer EEG measurements than the number of variables being inferred. Although neural population models can generate data resembling EEG, there are no clear mathematical solutions for the model parameters given the observation data. Indeed, there may be multiple solutions within the state space given the observation data. That is to say, there are multiple mathematical solutions to the same observation and not all are correct. Even though some of the solutions are mathematically achievable, it does not mean they are biologically explainable. Developing a technique that can track neurophysiological processes that is reasonable both mathematically and biologically will open new vistas for tracking and imaging brain states, as well as controlling them.

Regarding methods that have been used for time domain-based inference of neurophysiological processes given EEG data, the Kalman filter and its nonlinear variants have been applied (10,12,13,18). They address the problem of estimating the state of a discrete-time controlled process through stochastic difference equations (19). With the covariance matrix of the process and measurement noise, a Kalman filter balances the model and measurement noise to track the state vector of the model. However, it is a local search method that requires an initial guess of the expected state vector and its covariance. As a result, state vector estimates can deviate far from the true state vector if the initial guess is not close to the correct state vector. Furthermore, a solution to the Kalman gain and the covariance is needed. Since an arbitrary NMM can potentially be very high dimensional, the solution is not always easy to compute, and the initial guess is hard to determine because the neurophysiological variables cannot be measured in most cases. A solution that does not require an initial guess and is easy to apply given an arbitrarily complex model is needed.

Long Short-Term Memory (LSTM) is a recurrent neural network that can "remember" older information in a time-series sequence (20). Therefore, it is a potential approach to solve the problem of inferring underlying neurophysiological processes (i.e., the state vector and parameters of NMMs) from EEG data (21,22). In this paper, we compare the LSTM approach with a nonlinear variant of Kalman Filtering.

## 2. Methodology

### 2.1 Brain Modelling

Scalp and intracranial EEG record electrical activity signals captured on the scalp and on the surface of the brain, respectively. These signals primarily represent the macroscopic activity of cerebral cortex. A single EEG signal is recorded by a single electrode, and is a spatially and temporally smoothed version of neuronally generated electrical potentials (23). Intracranial EEG is invasive and difficult to place enough electrophysiological recording electrodes in a given cortical location to understand how the underlying neurophysiological processes give rise to the EEG. As a result, modelling is a way to bridge the EEG data and the underlying neurophysiological processes.

There are predominantly two approaches to build a mathematical model of the brain: the "bottom-up approach" and the "top-down approach" (24). The bottom-up approach forms a macroscopic model of the brain or a brain region by coupling a network of microscopic models of individual neurons (e.g., Hodgkin-Huxley neurons). To obtain the spatial scale relevant to modelling EEG signals, the large number of neurons involved in this kind of model, as well as the heterogeneity and non-local connectivity of neurons, makes this largely impractical for understanding which key population-level neurophysiological variables influence macroscopic dynamics.

We focus on one of the most common NMMs used to model cerebral cortex, the Jansen-Rit model (9,25). The Jansen-Rit model was developed based on the model proposed by Lopes Da Silva et al. (26), who proposed a two-population neural field model to study spontaneous EEG generation and alpha frequency rhythms, and reduced the complexity of the model by aggregating the activities within a single cortical column to perform analyses. Jansen et al. (25) constructed a similar model of pyramidal neurons to show that the mechanisms that guide the evolution of evoked potentials and spontaneous EEG signal generation are the same. Jansen and Rit (9) extended the model with a third population of local excitatory interneurons that provide feedback to the pyramidal neuron population. The latter model has three populations: pyramidal neurons, inhibitory interneurons and excitatory

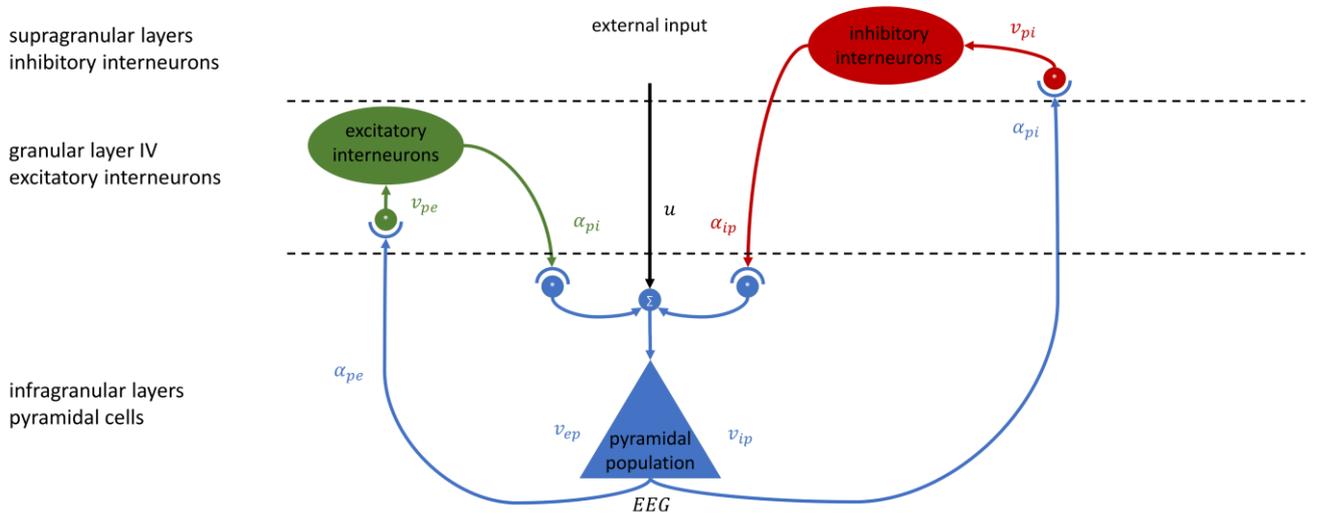

Figure 1. Jansen & Rit neural mass model and time series recordings and parameter values. Structure of the model. The model consists of three populations: pyramidal neurons, inhibitory interneurons and excitatory interneurons. The output of each population is a firing rate, which is transformed into changes in the mean membrane potentials of connected neural populations.

On the other hand, the top-down approach defines dynamical models based on phenomenological observations and emergent behaviour. This means that the mathematical model is built to model large-scale signals, such as EEG, through the interactions of neural populations, where the neural populations only coarsely model the average properties of the underlying detailed neuronal networks. Although it would be useful to model and infer the detailed microscopic neurophysiological processes, this is difficult to do as a result of the aforementioned experimental and observability issues as well as computational tractability. As a result, we focus here on inferring population-level neurophysiological processes using top-down models.

interneurons.

Figure 1 illustrates the model. The spatially averaged post-synaptic potential of population $m$ arising as a result of input from pre-synaptic population $n$ is expressed as

$$v_{mn}(t) = \alpha_{mn} \int_{-\infty}^{t} h_{mn}(t - t') \phi(v_n(t')) dt',$$

[1]

where $\alpha_{mn}$ is the population averaged synaptic connectivity strength. The post-synaptic response kernel is given by

$$h_{mn}(t) = \eta(t) \frac{t}{\tau_{mn}} \exp\left(-\frac{t}{\tau_{mn}}\right),$$

[2]

where $\eta(t)$ is the Heaviside step function and $\tau_{mn}$ is the population averaged synaptic time constant. Moreover, $\phi(v)$ is a sigmoid function that transforms the membrane potential to a firing rate given by

$$\phi(v) = \frac{1}{2}\left(\text{erf}\left(\frac{v-v_0}{\varsigma}\right)+1\right),$$  [3]

where $v_0$ is a threshold and $\varsigma$ is the slope of the sigmoid function, which is also the variance of firing thresholds of the population assuming the firing thresholds follow a Gaussian distribution. The mean membrane potential of population $m$ is the sum of the synaptic contributions,

$$v_m(t) = \sum_{n=1}^{N} v_{mn}(t).$$  [4]

The convolution in equation [1] can be also written as two coupled, first-order, ordinary differential equations (27),

$$\frac{dv_{mn}}{dt} = z_{mn}$$  [5]

$$\frac{dz_{mn}}{dt} = \frac{\alpha_{mn}}{\tau_{mn}}\phi_{mn} - \frac{2}{\tau_{mn}}z_{mn} - \frac{1}{\tau_{mn}^2}v_{mn},$$  [6]

where $\tau_{mn}$ is a lumped time constant, and $m$ and $n$ indicate the pre-synaptic and post-synaptic neural population, respectively.

Furthermore, we can express the neural mass model in matrix-vector notation so the operations can be expressed more compactly. The neural mass model can be expressed in matrix notation as

$$\dot{x}(t) = Ax(t) + B\vec{\phi}(Cx(t))$$  [7]

$$y(t) = Hx(t) + v(t),$$  [8]

where $H$ is the observation matrix, $v(t) \sim N(0, R)$ is the observation noise (10) and $y(t)$ is the membrane potential of the pyramidal population, which is considered to be the contributor to the generation of the EEG signal in our model. The matrix $A$ is defined by the membrane time constants,

$$A = \text{diag}(\Psi_1 \dots \Psi_n),$$  [9]

where

$$\Psi_n = \begin{bmatrix} 0 & 1 \\ -\frac{1}{\tau^2_{mn}} & -\frac{2}{\tau_{mn}} \end{bmatrix}.$$  [10]

The matrix $B$ is the synaptic gains from internal inputs, which has the form

$$B = \text{diag}(0\ a_1 \dots 0\ a_n).$$  [11]

The matrix $C$ is the adjacency matrix that defines the connectivity structure of the model,

$$C = \begin{bmatrix} 0 & 0 & \dots & 0 & 0 \\ c_{2,1} & 0 & & c_{2,N_x-1} & 0 \\ \vdots & & \ddots & & \vdots \\ 0 & 0 & & 0 & 0 \\ C_{N_x,1} & 0 & \dots & C_{N_x,N_x-1} & 0 \end{bmatrix},$$  [12]

which is a matrix of zeros and ones, where one indicates there is a connection between two populations and zero indicates there is no connection

In order to apply a nonlinear Kalman filter to this model to infer both the state vector and the parameters, we define a vector of parameters as $\theta = \begin{bmatrix} u\ \alpha_{pe}\ \alpha_{pi}\ \alpha_{ip}\ \alpha_{ep} \end{bmatrix}^T$. This set of parameters corresponds to the input $u$ to the model and the population averaged connection strengths between the three populations. Moreover, we assume the time constants of the model to be constant to simplify the estimation problem. The above parameters were combined with the state vector $x$ to form the augmented state vector,

$$\xi = [X^T \theta^T]^T.$$  [13]

The augmented state-space model is then

$$\xi_t = A_\theta \xi_{t-1} + B_\theta \phi(C_\theta \xi_{t-1}) + W_{t-1},$$  [14]

where $W_t$ is Gaussian noise.

### 2.2 Analytic Kalman Filter

Before defining the novel LSTM-based filter for state vector and parameter estimation, this section outlines the Kalman filter applied as a benchmark. In particular, a filter referred to as the Analytical Kalman filter (AKF) is applied. This filter is highly stable and accurate and was developed in prior work (2,10,27) that evolved from deriving the Kalman filter for general nonlinear NMMs using the specific sigmoidal non-linearity in equation (15). For the sake of brevity, we refer the reader to these prior works for greater mathematical insight. Moreover, links to code provided at the end of this paper give the exact specification of the implementation of the AKF. Here, a brief description of the main computations of the AKF are provided.

The aim of the AKF is to find the most likely posterior distribution of the augmented state given the previous measurements under Gaussian assumptions. Such a posterior distribution is characterised by its *a posteriori* state vector estimate and its covariance,

$$\hat{\xi}_t^+ = E[\xi_t \mid y_1, y_2, \cdots, y_t]$$  [16]

$$\hat{P}_t^+ = E\left[(\xi_t - \hat{\xi}_t^+)(\xi_t - \hat{\xi}_t^+)^\top\right]. \qquad [17]$$

The AKF proceeds in two steps: prediction and update. During prediction, the prior distribution (obtained from the previous estimate) is propagated through the model equations. This step provides the so called *a priori* estimate, which is also a Gaussian distribution with mean and covariance,

$$\hat{\xi}_t^- = E[\xi_t \mid y_1, y_2, \cdots, y_{t-1}] \qquad [18]$$

$$\hat{P}_t^- = E\left[(\xi_t - \hat{\xi}_t^-)(\xi_t - \hat{\xi}_t^-)^\top\right]. \qquad [19]$$

During update, the *a posteriori* state vector estimate is determined by correcting the *a priori* state vector estimate with recorded (EEG) data by

$$\hat{\xi}_t^+ = \hat{\xi}_t^- + K_t(y_t - H\hat{\xi}_t^-), \qquad [20]$$

where $K_t$ is the Kalman gain (18),

$$K_t = \frac{\hat{P}_t^- H^\top}{H\hat{P}_t^- H^\top + R}. \qquad [21]$$

The *a posteriori* state vector estimate covariance is then updated by using the Kalman gain,

$$\hat{P}_t^+ = (I - K_t H)\hat{P}_t^-. \qquad [22]$$

After each time step, the *a posteriori* estimate becomes the prior distribution for the next time step, and the process repeats for each new recorded data point. The AKF requires the *a posteriori* state vector estimate and its covariance to be initialised at time $t = 0$. The special features of the AKF are that the state vector estimate and its covariance, along with the Kalman gain, were derived using the fully non-linear Jansen-Rit model.

While the AKF framework can be applied to multichannel EEG data in a multivariate fashion, for simplicity this paper focuses on estimating the state vector and parameters of the Jansen-Rit model using a single channel of EEG data (analogous to the output observation/measurement of the Jansen-Rit model) as input to the filter.

### 2.3 A novel LSTM filter for state and parameter estimation

The Recurrent Neural Network (RNN) was introduced mainly to deal with time series data such as speech recognition and natural language processing (28). An ordinary feed forward neural network is only processes independent data points. However, when data points depend on previous data point, in the case of time series data, an RNN incorporates the dependency by providing feedback to the next timestep. In this way, the final output depends not only by the input but also on the outputs of previous neurons.

Long Short-Term Memory (LSTM) is a variation of the RNN that has improved its performance (20). Since gradient descent is applied to train neural networks, the gradient might explode or vanish after applying the sigmoid function over and over again, limiting to a certain number of discrete time steps to avoid this problem (29,30). However, the LSTM provides stability to bridge many more discrete time steps by enforcing constant error flow within special memory cells.

LSTM neural networks incorporate memory cells and gate units to convey useful information about previous states of the neural network to the current state. An LSTM layer consists of multiple recurrently connected memory blocks and three multiplicative gates, known as input, output and forget gates, that learn to open and close access to the error flow within the memory cell (20,31). For example, an input gate could use inputs from other memory cells to decide whether it should store information in its own memory cell.

An extension of the LSTM model is the bidirectional LSTM model that provide more accurate and improved results (32–34). One of the shortcomings of RNNs is that they can only learn from the context of the previous time steps and not anything after the current time step. Bidirectional LSTM can process data in both directions with separated hidden layers, which are then fed to the same output layer (35). Also, since the LSTM is free to access as much or as little of the data within the given time window as needed, predefining a specific time window for the model is not required (31).

Here, a bidirectional LSTM filter is constructed that takes as input simulated or real EEG and predicts the state vector and parameters of the Jansen-Rit model as well as simulated or real EEG. Similar to the AKF, while the framework that follows could potentially be applied to multichannel EEG data in a multivariate fashion, this paper, for simplicity, focuses on using the LSTM filter to estimate the state vector and parameters of the Jansen-Rit model for a single channel of EEG data.

The structure of the LSTM model is designed to achieve high performance while maintaining good time efficiency. Since the only non-linearity of the neural mass model is the firing rate function in equation (3), and to achieve high performance while predicting parameters given the observation, we would expect that at least two layers of artificial neural network are required. A grid search experiment was conducted to test the number of neurons in each layer. The selected structure was 128 and 32 neurons in layers 1 and 2, respectively.

### 2.4 Training Data

As real EEG data provides no ground truth about the neurophysiological variables that correspond to the state vector and parameters of the Jansen-Rit model, the LSTM filter is trained on simulation data was required. This allows the prediction of the state vector and parameters of the model when inputting either simulated or real EEG into the LSTM filter. It is important to generate a variety of training data with different patterns and a wide range of parameters so the model can learn the relationship between signal patterns and how the state and parameters change. Furthermore, it is also crucial to generate data containing oscillations, since there are likely to be many solutions for a constant signal, or fixed point. This is because when there is no oscillation, different combinations of changes in either the external input, input from the inhibitory or excitatory interneurons may result in the same observation signal. However, changes of the parameters could affect the ranges of external input that generate oscillations, which makes it difficult to find the desired range of external input when there can be different combinations of parameters. This is because the time constants affect the transformation from firing rate to post-synaptic membrane potential thus affecting the ranges over which the population produces oscillations. Thus, an effective method to generate oscillatory data with a wide range of parameters is required.

Different combinations of inhibitory and excitatory time constants can generate data with different waveforms, frequencies and amplitudes. Normally, the ranges of both excitatory and inhibitory time constants are considered between 10-60 ms (36). In addition, external input also plays an important role in data generation. External input to the model is also affected by the time constant, such that the range of the external input that can produce oscillations is affected by different combinations of time constants, but the exact range for each combination is unknown. To determine the range of external input and whether the given combination of time constants is able to generate oscillation, an automatic data generation process is designed.

Statistical hypothesis testing is used to determine whether an oscillation exists in a given time-series recording. Two tests are used. If the simulated EEG signal is noise-like, then it will have a Gaussian distribution so the Anderson-Darling test (37) is used to test if the data is not Gaussian distributed, which means there could be an oscillation. The other test is the Ljung-Box test (38), which tests whether any autocorrelations of the time series are different from zero. Either of the tests can help determine whether the generated data contains an oscillation and, since it is more important to ensure there are no false positives (data is Gaussian distributed or data has non-zero correlation but we reject this hypothesis) to avoid data being generated without including oscillations. We set the significance level to $\alpha = 10^{-4}$.

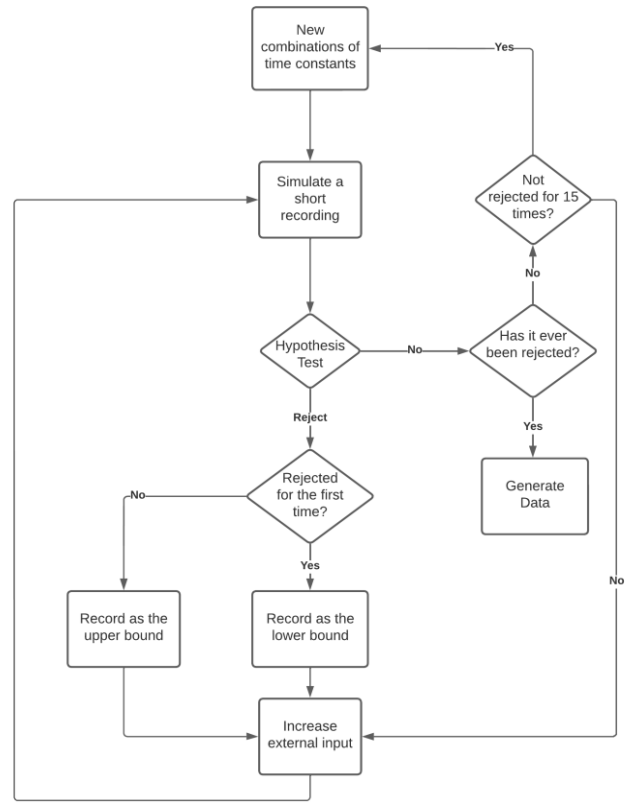

*Figure 2. Flow chart of the generation of training and testing data. The same flow chart is used on different NMMs (the Jansen-Rit model is used in this paper). The amount of the increased external input is not fixed, as the upper bound could be high when the excitatory time constant is high. Thus, the amount is low when the input is low, and is higher when it is high.*

A flow chart used to generate the training data is shown in Figure 2. We generate a short recording first to check whether this combination of time constants and external input is able to generate data containing oscillations. If not, then the external input is increased until it does. If it still does not produce oscillations for 15 consecutive increases of the input, the current combination of time constants is considered to not be able to produce the desired data and so the next combination is used. If the generated data includes oscillations then the upper and lower bounds of the external input are recorded, and training data is generated with the recorded parameters.

The size of the dataset is as large as possible within the size of physical memory, which means that the gap between each combination is kept relatively small and as many examples as possible are provided for training (39). The generated data was shuffled randomly and divided into training, validation and testing data with ratio of 80:10:10.

To improve the performance of the model with different amplitudes, the dataset is standardised. As in many cases with either real or simulated data, the raw amplitude of the EEG data may be on arbitrary scales. To avoid potential issues due to variations in amplitude scale or values out of the training range, The mean and standard error were computed for each feature and target variable, and standardisation is implemented for each of them using $\frac{x-\mu}{\sigma}$. The other reason for doing this is to ensure the loss is reduced for all variables without any bias, as different amplitudes might cause the loss of some of the target variables to be weighted more than others. For example, if one variable has a higher amplitude than the other, its error is higher and thus will be reduced more.

*2.5 Loss Function*

The training of an artificial neural network aims to minimise the loss function over all training data. By default, the loss function is the mean squared error of all target variables, which are the state variables and parameters in our case. However, it is not enough for the model error to be minimised only on the training data, as we would like the LSTM model to learn whether the predicted state variables and parameters can represent the observations. More importantly, it should also learn to behave like the NMM. Given an observation signal, the trained LSTM model should be able to predict the state vector and parameters that can recreate the observation signal using the NMM. Since the generation of the training data is based on setting the discrete parameters, even though we can lower the step within each parameter, it is still possible that the model cannot learn the mathematical relationship between all state variables and parameters. Thus, we link the LSTM model with the NMM.

To link the LSTM model with the NMM, we customise the loss function to allow the LSTM model to know the mathematical relationship between the observation and the state. Since the LSTM model is able to produce the state for the next timestep $t$ given the state of the current time step $t-1$, it is possible to generate all $t$ states given the state at the current timestep. By comparing the state at time $t$ with the state predicted by the LSTM model, we can know the error between the LSTM model prediction and the neural mass model prediction. If we can link the neural mass model to the loss function, the LSTM model can learn to minimise both the error of the LSTM predicted values and the neural mass model predicted values.

The squared error at a single timestep is defined as

$$\text{Squared Error} = \left[ \left( \xi_t - \hat{\xi}_t \right)^2, \left( y_t - H\hat{\xi}_t \right)^2 \right], \qquad [22]$$

where $\xi$ is the augmented state, $y$ is the observation of the training data and $\hat{\xi}$ and $H\hat{\xi}_t$ are the augmented state and observation predicted by the LSTM model, respectively. The Squared Error term compares the predictions with the true values. By minimising the error, the predictions of both states and measurements will be closer to the truth.

We then add to the loss function the model error, which is the error between the state of the LSTM model and the state generated by the neural mass model (40,41),

$$\text{Model Error} = \left[ \left( \hat{\xi}_t - \left( A_\theta \hat{\xi}_{t-1} + B_\theta \phi \left( C_\theta \hat{\xi}_{t-1} \right) \right) \right)^2, 0 \right] \quad [23]$$

With the augmented state-space model, the estimated augmented state vector at the current time step $\hat{\xi}_{t-1}$ can be converted to the next time step. Minimising the squared error between $\hat{\xi}_t$ and the converted state vector $A_\theta \hat{\xi}_{t-1} + B_\theta \phi \left( C_\theta \hat{\xi}_{t-1} \right)$ links the estimation to the neural mass model, since the training has to minimise the error between its own prediction and the state generated by the NMM.

Note that, although $B_\theta$, $C_\theta$ and $\phi$ are considered to be constant, the matrix $A$ can vary depending on the time constants; i.e., the $A$ has to change over time. In the customised loss function, $A$ is calculated at each timestep so the change in state and parameters can be reflected over time.

The last thing aspect of the loss function is the rate of change of the connectivity strength parameters, as they are considered to be slowly changing parameters compared to the membrane potentials and simulated EEG. It is possible to add the standard deviation of the parameters to the loss function to limit the amounts that they change. However, the parameters need to be adjusted rapidly when they differ substantially from the neural mass model. Thus, the standard deviation is combined with the model error as

$$Std = s(\alpha) \left[ \left( \hat{\xi}_t - \left( A_\theta \hat{\xi}_{t-1} + B_\theta \phi \left( C_\theta \hat{\xi}_{t-1} \right) \right) \right)^2, 0 \right] * k$$
$$[24]$$

$$s(\alpha) = [0, \ldots 0, \text{std}(\alpha), 0]^T, \qquad [25]$$

where $\alpha$ is the vector of the four connectivity strengths parameters and $k$ is an adjustable weight that is set to 0.1 as default.

The final loss is the summation of equations [22], [23] and [24].

*2.6 Testing on Simulated EEG Data*

On top of the testing data split from the training dataset, another testing dataset with the same range of parameters was generated, but the steps of $\tau_e$ and $\tau_i$ are set to be smaller. This dataset avoids using the exact parameters of the original training dataset.

We tested both the AKF and the LSTM model on the same dataset. To evaluate the impact of incorrect initialisation of the AKF, two settings were used. The first setting was correctly initialised Kalman filters that were initialised to be exactly the same as the parameter initial values. The second setting was default parameters, where the excitatory and inhibitory time constants were 0.02 s and 0.01 s, respectively, which corresponds to the parameter values that generate a strong alpha-like (8-12 Hz) rhythm (9).

To ensure all state vectors and parameters are on the same amplitude scale numerically, the testing results are standardised. We compute the Root Mean Squared Error (RMSE) between the truth and the prediction for each state variable and parameter, and find the difference between the AKF and the LSTM model result. The RMSE for each parameter is

$$\text{RMSE} = \sqrt{\sum_{i=1}^{n} \left( \frac{x_i - \mu}{\sigma} - \frac{\hat{x}_i - \mu}{\sigma} \right)^2 / n}, \qquad [26]$$

where $\mu$ and $\sigma$ are the mean and standard deviation of the testing dataset for each parameter, respectively, and $x_i$ and $\hat{x}_i$ are the truth and prediction, respectively. Equation (26) is used for each estimation method to extract the raw RMSE and the values are compared across methods.

Testing the robustness of the model is also of great importance, since typically there will be strong noise in real EEG data. As the standard deviation of the testing observation data is around 40, we add Gaussian noise with a mean of 0 and a standard deviation of 4, representing 10% Gaussian noise added to the testing data. Both the LSTM model and the perfectly initialised Kalman Filter are tested using the noisy testing data.

We also test the LSTM model on randomly generated data involving time-dependent parameter changes. The time constants are randomly chosen from the range 0.01-0.06 s, and the model is simulated with the selected time contents held fixed for 5 s before the next combination of time constants. Note that there is a 5 s buffer between each 5 s fixed time constant simulation segment, where we connect the parameter values between segments with a straight line to simulate transitions between different brain states, where the parameters are not constant. We assume the parameters are slowly-changing, so a straight line between different stable states would be enough for simulation purposes. Since the training was done on constant parameters with noise only, testing on randomly generated data is needed to know whether the LSTM filter can work in a scenario where parameters are time varying.

### 2.7 Testing on Real EEG Data

Apart from simulation data, we also use intracranial EEG (iEEG) data collected from a patient with epilepsy chosen from data recorded continuously from 15 patients for up to three years (42). The patients were implanted with 16 intracranial electrodes around their seizure onset areas with sampling rate 400 Hz. The electrode array was wirelessly connected to a portable device. Seizures were automatically detected by the advisory device and were reviewed by clinical experts. We denote the iEEG data from this study as "real data".

To demonstrate the practical utility of the LSTM filter, it is applied to a 1 h intracranial EEG recording containing an epileptic seizure to show whether or not we can observe a transition between normal and ictal brain states. Although the actual values of the state variables and parameters are not known, there is an obvious brain state change with seizure onset, so we can see if the LSTM model may be used to detect or even predict a seizure. Since normally iEEG data becomes more rhythmic when it enters a seizure compared to normal activity, we expect the parameter estimates to become stable after entering the seizure state.

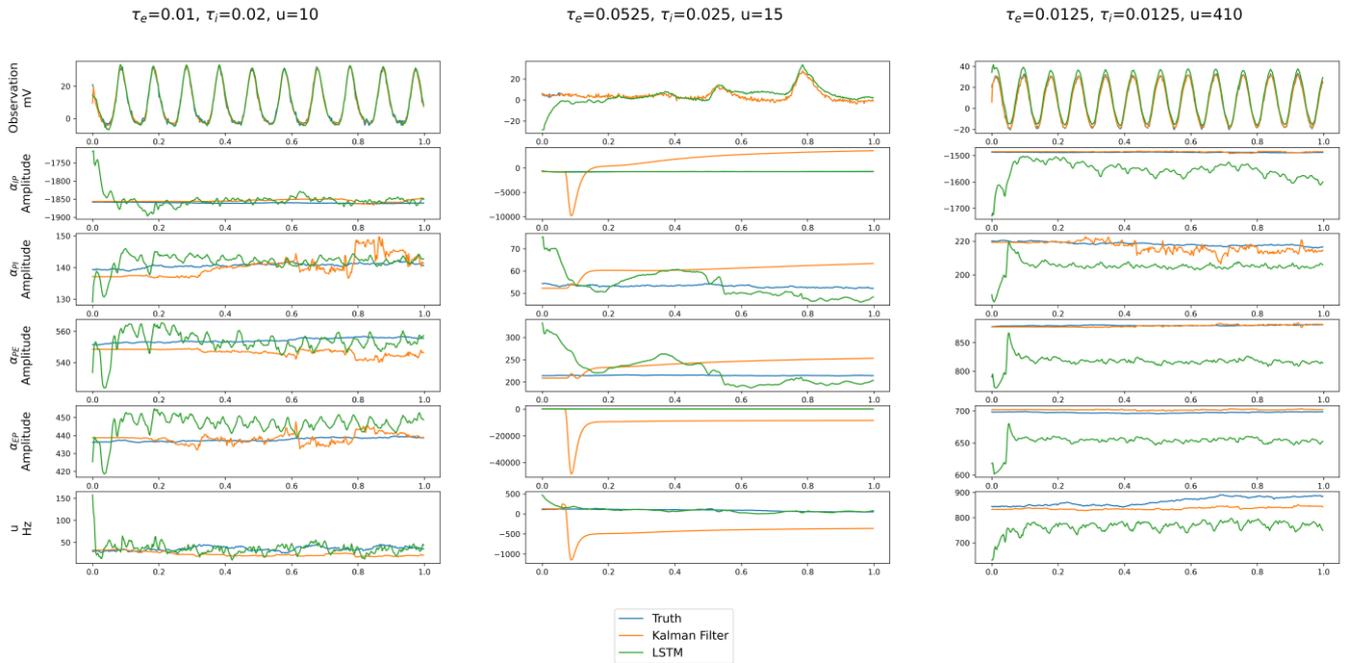

*Figure 3. Examples of time series recordings and predictions using Kalman filters and LSTM models. Different settings (columns) are given to the NMM to generate artificial EEG recordings. Observations and parameters are displayed to show the correct parameters (truth) and those estimated by the Kalman filters and the LSTM models.*

## 3. Results

We have chosen some typical time series to show the different parameter estimation results between the perfectly initialised AKF and the LSTM model. As shown in Figure 3, the first column is the default condition where excitatory and inhibitory time constants are 0.01 s and 0.02 s, respectively. The second and the third columns are examples where the LSTM performs better and where the AKF performs better, respectively. The simulated EEG signals can be fit closely for all scenarios. However, the fits of the parameters are very different. The LSTM model tends to converge rapidly in about 0.1 s, while the AKF takes more time to find the optimal solution. The LSTM model also appears to fluctuate more compared to the AKF, normally because the LSTM model has to compute using the provided input which is the observation signal. When the input has oscillations, the LSTM model can only minimise the changes, but it cannot make it constant like the truth. However, the results of the LSTM model are generally within a reasonable range with predictions close to the truth. The perfectly initialised AKF tracks the parameters very well, but the filter drifts far away from the correct solution for some cases, as shown in the second column.

Figure 4 shows comparisons between the proposed LSTM method against the AKF. The two time constants were selected as the axes of the plots since the time constants have a fixed range (0.01 s to 0.06 s) compared to the external input, which has a more flexible range depending on the time constants. The median of the RMSE is computed over different external inputs, as sometimes the AKF results in errors when it cannot find the nearest positive definite covariance matrix which leads to extremely large RMSE. The covariance matrix must remain positive definite to ensure inversion is possible, so a nearest positive definite matrix has to be found when it is not. Due to the standardisation of the parameters in the RMSE calculation, most of the raw RMSE results are within the range 0-1. Hence, the minimum and the maximum of the colour bars are set to 0 and 1, respectively. The minimum and the maximum of the colour bars for the RMSE difference are set to -1 and 1, respectively.

As seen in Figure 4 (a), the raw RMSE are mostly low for the LSTM model with a few exceptions. The performance is better in the centre of the parameter space compared to the edges. The centre of the space is close to the default parameter point ($\tau_e = 0.01, \tau_i = 0.02$) for the Jansen-Rit model. The regions of low performance (red area) are because very little or no oscillations are detected in the observation signal. The perfectly initialised AKF has a nearly perfect performance for lower values of the excitatory time constant (the upper parts of the plots), but the result is worse when the excitatory time constant increases. On the other hand, the fixed AKF initialised to the default parameter point only performs well around the region of the default setting.

Figure 4 (b) show the differences between the LSTM filter and the two initialisation cases of the AKF. The red zone indicates that the LSTM model has a higher RMSE than the AKF. For the LSTM and the perfectly initialised AKF, despite the lower part of the graph where the LSTM model has a much lower RMSE compared to the AKF, the LSTM model usually has a slightly higher RMSE. Compared with the fixed Kalman Filter, the LSTM model usually shows much better performance. Thus, we can conclude that the proposed LSTM model can be considered as a generally better tool to estimate the EEG data without knowledge of the initial state variables and parameters.

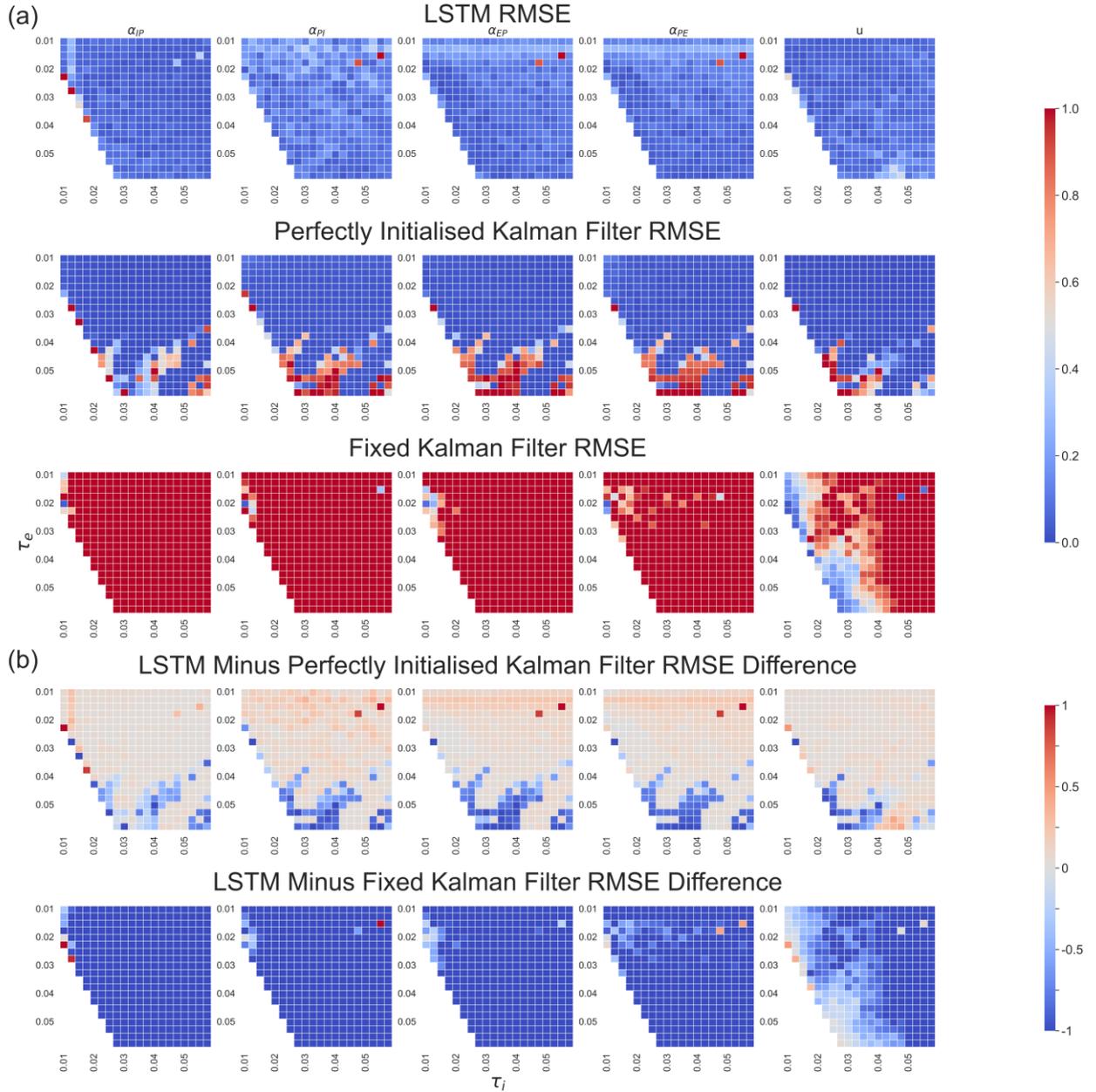

*Figure 4. Heatmaps of the LSTM and Kalman Filter errors. (a) Raw RMSE between the prediction and the truth. The test was run on the LSTM model (top), the perfectly initialised Kalman Filter (middle) and the fixed Kalman Filter (bottom) whose initialisation was set as default. The colours indicate the amount of error ranging from 0 (blue) to 1 (red). (b) The RMSE difference between the LSTM and the Kalman Filters. The difference was compared between the LSTM and the perfectly initialised Kalman Filter (top), and between the LSTM and the fixed Kalman Filter (bottom). The comparison was done by using the RMSE of the LSTM minus the RMSE of the Kalman Filter, which means a positive number indicates the LSTM model having a higher error. The colours indicate the RMSE difference ranging from -1 (blue) to 1 (red), with no difference shown as grey.*

Figure 5 shows the performance with Gaussian noise added to all recordings with 0 mean and 10% of the standard deviation of the original data. We can see that the performance of the LSTM model did not change much, but the AKF did have more areas where the performance was worse compared to no noise. Furthermore, we can also observe from Figure 5 (b) that there are more blue areas compared to those in Figure 4 (b), showing that the LSTM model is more robust when the data is noisier.

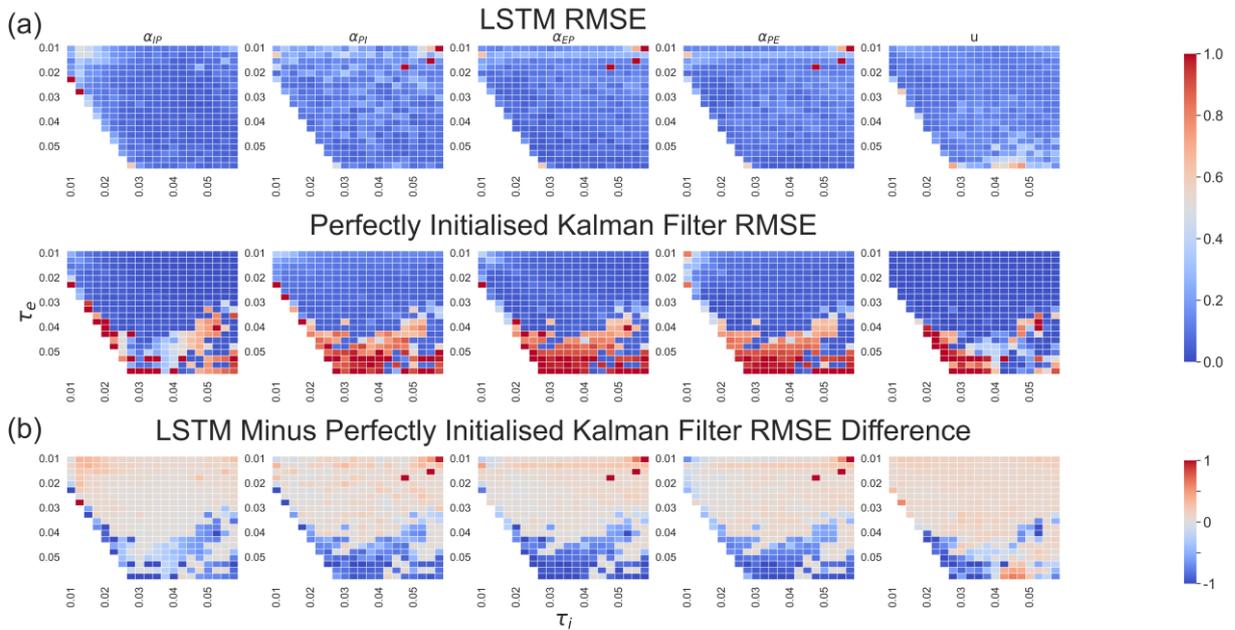

*Figure 5. Heatmaps of the LSTM and the Kalman Filter after adding Gaussian noise. The same method was applied and the perfectly initialised Kalman Filter was tested. (a) Raw RMSE between the prediction and the truth. The test was run on the LSTM model (top) and the perfectly initialised Kalman Filter (bottom). The colours indicate the amount of error ranging from 0 (blue) to 1 (red). (b) The RMSE Difference between the LSTM and the Kalman Filter. The comparison was done by using the RMSE of the LSTM minus the RMSE of the Kalman Filter, which means a positive number indicates the LSTM model having a higher error. The colours indicate the RMSE difference ranging from -1 (blue) to 1 (red), with no difference shown as grey.*

To show the performance of the LSTM filter in response to time-varying parameters, we also test using randomly generated recording with changing parameters as presented in Figure 6. The AKF is initialised with the initial parameters of the simulation data. Its parameter estimates are stable at the beginning of the recording when the true parameters are stable, but its parameter estimates diverge when the true parameters change. Although AKF can follow the observation at the beginning (before 20 s), it does not follow the parameter space correctly after this period. The LSTM model shows a much better fit to both the observation and the parameters. There are some times when the parameters are not followed precisely but, overall, the LSTM model tracks within a reasonable range. As a result, the figure shows that the LSTM model is able to track time-varying changes in the true parameter values that might be observed when the brain transitions between different dynamical states.

The final test involves using real intracranial EEG data (42) containing an epileptic seizure to show whether the LSTM filter can detect brain state changes. In Figure 7, the seizure start and end times are labelled with vertical red bars, though they are difficult to differentiate in the plot of the full hour of data. In Figure 7 (a), which shows the estimation results for one hour of data, we can see all parameters fluctuate frequently and in a large range as well, which reflects the noisy EEG recording with no stable state to track. However, upon entering the seizure state, the changes in the parameters were limited as seen from Figure 7 (b), where the parameters are more stable during the seizure, and fluctuate more before and after the seizure. We can also see that from Figure 7 (b), where we zoom into the minute when the seizure was about to begin, the parameters transition between different values. This shows some promise that the LSTM model may be able to detect epileptic seizure transitions.

## 4. Discussion

A novel LSTM filter has been presented to provide efficient neurophysiological variable estimates derived from EEG data, which does not require initialisation of the filter and can track dynamic changes in brain states. Moreover, the approach is sufficiently general that it can potentially be applied to other neural mass models.

Although the overall performance of the LSTM model is better or at least similar to the perfectly initialised AKF, there are some noticeable drawbacks. In most cases, we can see that the performance of the estimated external input is usually the worst among all state variables and parameters, especially in the heatmaps shown in Figures 4 and 5. This is because there are few mathematical constraints regarding the external inputs, as it acts as a way to adjust DC shift that is sometimes not achievable by the membrane potential linked to the pyramidal population. Other parameters that are more constrained mathematically by the relationship between the neural mass model equations and the EEG output of the model have better estimation performance. On the other hand, converting the firing rate of the external input to the membrane potential via the post-synaptic potential kernel depends on the excitatory time constant, so the external input has a dynamic range to vary instead of a fixed range like the time constant, which makes it harder to track. However, the situation improves when the data becomes noisier as the LSTM is still robust compared to the AKF.

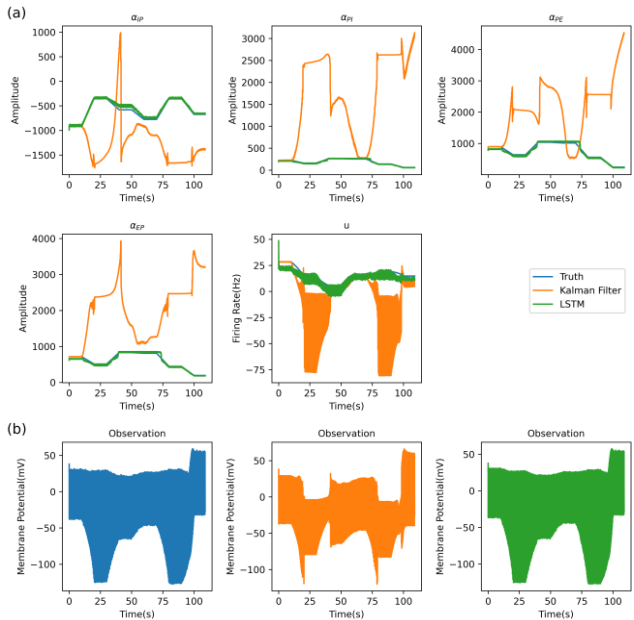

*Figure 6. Test on a randomly generated recording. (a) True parameters and the prediction made by the LSTM and the AKF. The AKF is perfectly initialised at the beginning, but starts to lose track as the change of parameters happens, while the LSTM keeps mostly on track. (b) Observation signals. The LSTM model follows the truth much better than the AKF.*

Time efficiency is also one of the important features of the artificial neural network. The AKF only considers one previous timestep, but the LSTM model considers 400 timesteps. This means the computation of the LSTM model is more complex. We have tested the result when both methods are computed via CPU, the AKF takes about 516 seconds to run on a one-hour recording, while the LSTM model takes about three times longer: about 1720 seconds. With the utilisation of GPU, the LSTM model is significantly faster, and is able to run on the same recording in 14 seconds. This suggests the LSTM filter can be scaled up for larger more complex neural mass models for more detailed inference based imaging tasks (2), while still providing tractable run times.

Another advantage of the LSTM filter is that it can potentially be trained on a very specific region of the parameter space, in particular only within biologically realistic regions of the model parameter space. This means it will be much more likely to produce estimates in such a region also. The AKF on the other hand is dependent on numerical stability of the neural mass model associated with the sampling rate being used, and it also is not necessarily constrained to producing estimates outside of biologically plausible regions of the parameter space. Thus, in some cases the AKF can produce unrealistic estimates. In this paper, the training data generated by the neural mass model sometimes did not comply with biological realism. For example, the firing rate sometimes achieved rates as high as ten thousand spikes per second, which is not biologically realistic. This is because the purpose of the experiment was to show whether the LSTM filter is able to track the state vector and parameters regardless of the range of the parameters in general. Nevertheless, it would be straightforward to train the LSTM filter on completely biologically realistic regions of parameter space, in order to only produce biologically realistic estimates. Since there is only one input dimension, the observation in our training data, we would like to consider whether adding more features would benefit. We could extract the band power from one second prior to the current timestep to represent the change in the strength over different frequency bands. Since the frequency changes might be minor, especially in the low frequency area, we would like to compute a narrow bandwidth within low frequencies and wide bandwidth for the high frequencies (43). However, this feature extraction process takes a long time leading to time efficiency issues.

The LSTM filter presented here is accessible through the link that is provided in the Code Availability section. The package has been prepared such that the LSTM filter can be applied in a univariate fashion to multichannel EEG, intracranial EEG or EEG/MEG source imaging data to derive single channel neurophysiological variable estimates across all channels. Future work will consider the multivariate multichannel estimate problem.

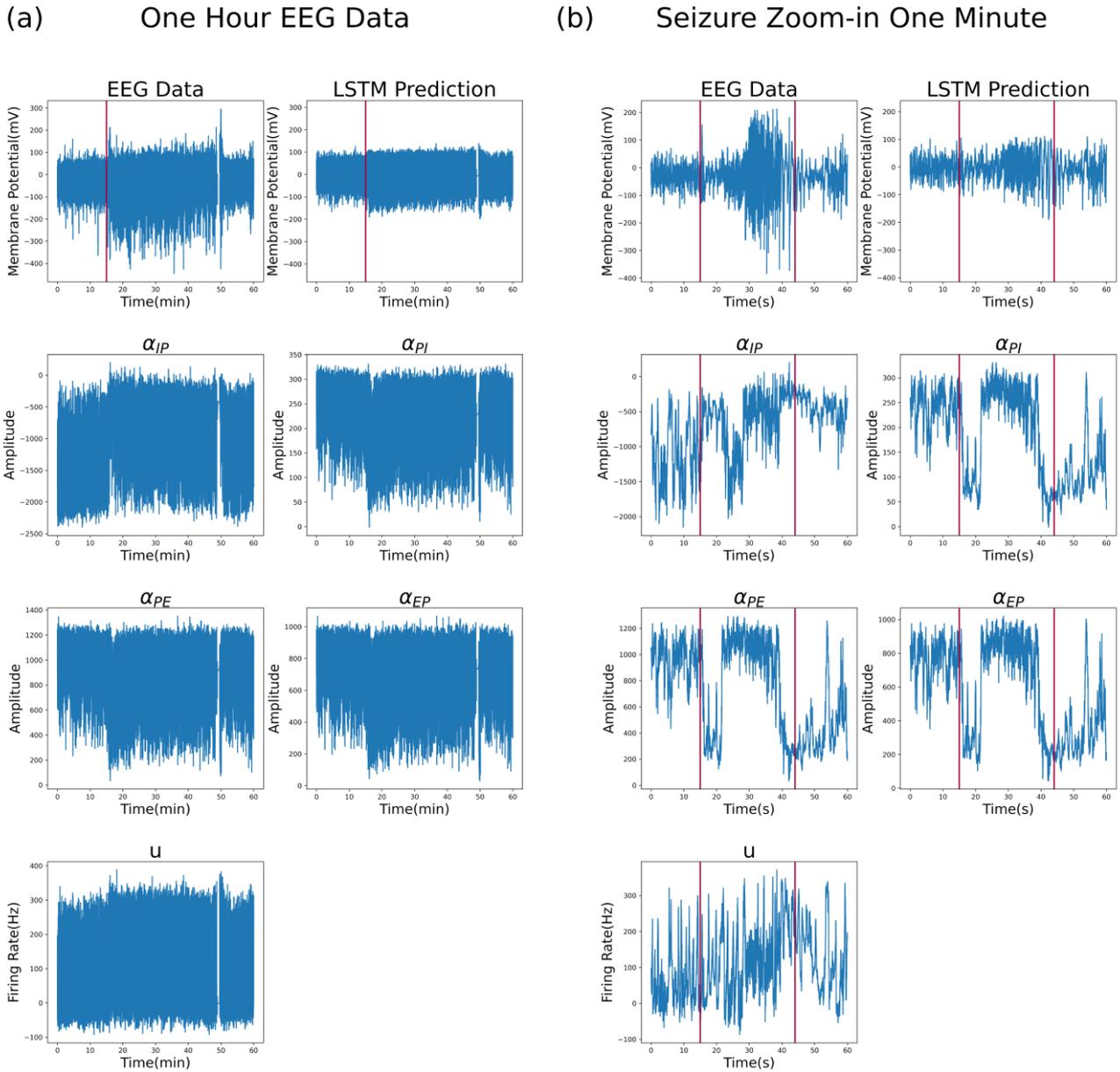

*Figure 7. Test on real data with a seizure. (a) A one-hour recording containing an epileptic seizure. Seizure is indicated as a vertical red bar. (b) Zoomed in for the minute where the seizure happened. The vertical red bars indicate the start and the end of the seizure.*

## 5. Conclusion

The proposed methodology has demonstrated competitive accuracy, high time efficiency, and the potential to be applied to real world scenarios such as epileptic seizure detection. The result has shown the accuracy of the proposed approach is comparable with a perfectly initialised AKF, and is much better than the AKF that is initialised with default parameters. In a parameter-changing environment, the LSTM filter is able to track the changing parameters much better than the AKF even if the AKF is perfectly initialised. With GPU acceleration, the time efficiency is also greatly improved, with the time cost reduced by 96%. Finally, after testing on real data with an epileptic seizure, the LSTM filter can detect instantaneous transitions in brain states and, therefore, holds promise as a potential application for detecting or predicting seizures. Further research will be focused on applying the proposed method to different scenarios to infer, image and understand the neurophysiological processes underlying different kinds of electrophysiological and electromagnetic brain recordings.

## 6. Code Availability

The data generation, model training and random recording testing codes are available at https://github.com/yueyang6/brain_state_reconstruction.